\documentclass[jou,apacite]{apa6}
\usepackage[utf8]{inputenc}
\usepackage{url}

\title{Political Footprints: Political Discourse Analysis using Pre-Trained Word Vectors}

\author{Christophe Bruchansky}
\affiliation{christophe@plural.world}

\abstract{In this paper, we discuss how machine learning could be used to produce a systematic and more objective political discourse analysis. Political footprints are vector space models (VSMs) applied to political discourse. Each of their vectors represents a word, and is produced by training the English lexicon on large text corpora. This paper presents a simple implementation of political footprints, some heuristics on how to use them, and their application to four cases: the U.N. Kyoto Protocol and Paris Agreement, and two U.S. presidential elections. The reader will be offered a number of reasons to believe that political footprints produce meaningful results, along with some suggestions on how to improve their implementation.}

\rightheader{Political Footprints}
\leftheader{Christophe Bruchansky}

\begin{document}
\maketitle    


\section{Context and Methodology}
Vector space models (VSMs) represent words in a continuous vector space where semantically similar words are mapped to nearby points~\cite{vsm}. VSMs are produced by algorithms that can analyze large corpora of text and determine how likely two words are to appear in the same passage (word "co-occurrence"): the more often two words appear together, the closer these algorithms will place their vectors. The resulting vector space models are not only statistically significant, but also have a semantic value. This is due to the distributional hypothesis stating that words appearing in the same context share semantic meaning~\cite{semantic2}. \par
Vector space models allow machines to classify documents by meaning, understand natural language, and, in our case, create "semantic word clouds". They provide "a bird’s-eye view of different text sources, including text summaries and their source material. This enables users to explore a text source like a geographical map”~\cite{map}. \par
Publicly available vector space models have made their appearance in recent years (“pre-trained” vector space models). Examples are word2vec models from Google~\cite{google}, GloVe models from Stanford University~\cite{glove} and FastText models from Facebook~\cite{fasttext}. All have further accelerated the adoption of the technology.\par
Cultural and political sciences are a natural fit for VSMs, and many research studies have started using the technology to analyze political opinions. A large proportion focus on social media, allowing, for instance, the categorization of election-related tweets; others focus on the political discourse itself, such as argument-based analysis~\cite{argument} and semantic word clouds~\cite{everything}. This paper belongs to the latter category. \par
A political footprint is a vector-based representation of a political discourse in which each vector represents a word. Political footprints are computed using machine learning technologies, which allows for a systematic and more objective political analysis.

\begin{figure}[h]
\caption{Donald Trump’s political footprint during 2016 U.S. general election debates, with the closest words to “people” highlighted in orange - principal component analysis projection from Tensorboard.}
\centering
\includegraphics[width=0.47\textwidth]{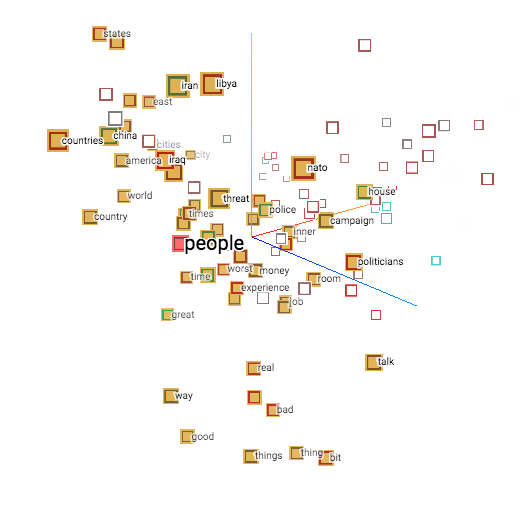}
\label{fig:trump-people}
\end{figure}

Because political footprints compute semantic similarity based on large corpora of text, they lead to political discourse analysis that relies less on the researcher’s political knowledge or set of beliefs. They are, however, very much dependent on the corpus they were trained with (Wikipedia, Google News, etc.), and more generally on the cultural context any political discourse originates in.\par
Political footprints focus exclusively on what a statement or speaker says. They are, in this respect, very different from other popular word clouds that depict news and social media trends: the emphasis is on what a speaker has in his or her control.\par
Our purpose in this paper is to establish a proof of concept: to use existing technologies and implement a simple version of political footprints in order to see if the results are in any way meaningful. It is based on the following technologies:
\begin{itemize}
  \item IBM Watson~\cite{watson}: returns a list of entities and keywords included in a text, with a relevance, sentiment, and emotion score for each of them;
  \item GloVe from Stanford University~\cite{glove}: package featuring word vectors trained using Wikipedia (2014) and Gigaword 5 (2011);
    \item  TensorFlow and TensorBoard~\cite{tensorflow}: open-source software library for machine intelligence, its TensorBoard interface is particularly useful for VSM quick visualisation.
\end{itemize}

The sections below describe each of the steps taken to produce political footprints. The first is to process raw data, the second is to identify key terms, the third is to compute political footprints, and the last is to interpret them using heuristics. All scripts and data are available on Github~\cite{github}.

\section{Raw data}
Four case studies were chosen to prove our concept: the Kyoto Protocol, the Paris Agreement, and the 2008 and 2016 U.S. presidential elections. The Kyoto Protocol and Paris Agreement were taken from the United Nations website~\cite{un}, and the U.S. elections televised debates from the American Presidency Project~\cite{president}, a "nonprofit and nonpartisan source of presidential documents" hosted at the University of California.\par
They have been chosen, firstly, because most readers will be familiar enough with the topics to compare the results with their own intuition, and, secondly, because they are accessible to anyone wishing to consult the data.\par
In the case of political elections, televised debates were chosen over rallies and other public declarations because they are assumed to be representative of the U.S. political landscape: processes defining which candidates were invited to the televised debates, how long they could speak, and on what matters are assumed to be objective enough for our purpose. We can thus easily apply the same method to any election. However, we are aware that no debate can perfectly reflect a political landscape. In the case of the U.S. elections, this assumption leads us to exclude independent and third party candidates, which is arguably regrettable. \par
The goal of this study is to focus on a political discourse, not on its context or how it has been received. We have thus removed any audience reaction from the transcripts (i.e.  “applause”, “laugh”, etc.) and all questions from the moderators. We lose a certain amount of important information in doing so, which means it is sometimes difficult to understand what a candidate’s answer was about, but this is the price we pay when we only take into account the candidates’ own words.
\section{Key terms identification}
Our choice to use IBM Watson is a convenient one: it allows us to run our analysis on any personal computer, and with fast results. It comes with a cost: there is not much control over or explanation of how the terms are selected and weighted. For instance, IBM Watson Natural Language Understanding generates several files per text, including one for its entities and one for its keywords. It’s not clear how the two lists are created, and it is assumed that entities are more relevant than keywords since they are more structured objects (entities have types and subtypes). If a term exists in both files, our script only keep the entity version.\par
As the reader will later realize, IBM Watson’s emotion detection is not always totally reliable. A better solution could be to use an emotion detection mechanism that can adapt to political language~\cite{emotion}. In any case, it is important to keep in mind that an emotion attached to a word is not necessary targeting that word. An angry feeling detected when using the word “people” does not mean that the discourse is necessary expressing anger towards people, but that speaking about people generates anger. In the case of Hillary Clinton's political footprint, we have also noticed how much linguistic features such as sarcasm remain problematic for tools like IBM Watson. \par
No other commercial (i.e. Google CloudPlatform) or open source solutions (i.e. NLTK or Stanford CoreNLP) have been tested as part of this paper. Words selected by the default IBM Watson API were for the most part corresponding to our intuition. They were considered good enough for our proof of concept, in the sense that if using a generic tool such as IBM Watson can provide meaningful data, more sophisticated implementations could only improve our results. \par

\section{Political footprints}
We have chosen GloVe6B for our pre-trained vector space model, which is based on Wikipedia 2014 and Gigaword 5. We took this decision due to GloVe's public availability, its popularity among researchers, and a number of convenient features, such as being able to quickly run our scripts using a 50-dimension space, before extending it to 300 dimensions. A VSM that defines each word using 50 variables is good enough for early tests, but it will not capture as many nuances as one based on 300 variables.\par
\begin{figure}[h]
\caption{Bernie Sanders’ key topics detected during 2016 U.S. election televised debates (PCA projection from Tensorboard).}
\centering
\includegraphics[width=0.47\textwidth]{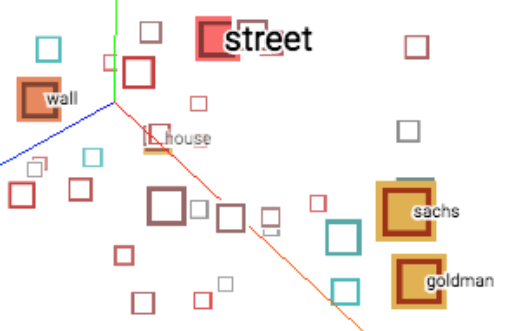}
\label{fig:wall-street-2016}
\end{figure}

We have, however, encountered several issues. The first is that this model is word based and does not include compounds nouns such as “Wall Street” or “New York Times” (see figure \ref{fig:wall-street-2016}). This was not so much of an issue in our case study since we were familiar with the data. We could easily guess the compound nouns: we knew that “street” was a relevant word in Bernie Sanders’s political footprint because of Wall Street. However, this requires a degree of guesswork and interpretation from the researcher that could be avoided if the pre-trained vector space model supported compound nouns.\par

A second issue is the date of the reference corpora (2014 for Wikipedia and 2010 for Gigaword 5) compared to the times of the U.S. elections (2008, 2016). We could only assume that the words’ meaning and usage remained unchanged during that period. An improvement would be to have access to yearly updates of pre-trained vector space models. However, this would create new issues, such as how to compare texts written in different years if we do not base our analysis on a single reference model. \par
Another problem is that GloVe's pre-trained vector space models are currently only available in English. A multi-language alternative is FastText from Facebook, which is compatible with our scripts. It was, however, only few months old at the time of our research, and we decided to stick with a more established GloVe model for our proof of concept.\par
Let us now look at how we can put political footprints to practical use.
\subsection{Heuristic 1: main themes of a discourse and what they mean}

The first heuristic could be described as a clustering technique in which we leverage the relevance score obtained from IBM Watson: we take the most relevant words from a discourse, and select the closest words in the vector space model for each one. We obtain a series of clusters each centered on a relevant term.\par

\begin{figure}[h]
\caption{Closest words to "climate" in the Kyoto protocol (PCA projection from Tensorboard). }
\centering
\includegraphics[width=0.47\textwidth]{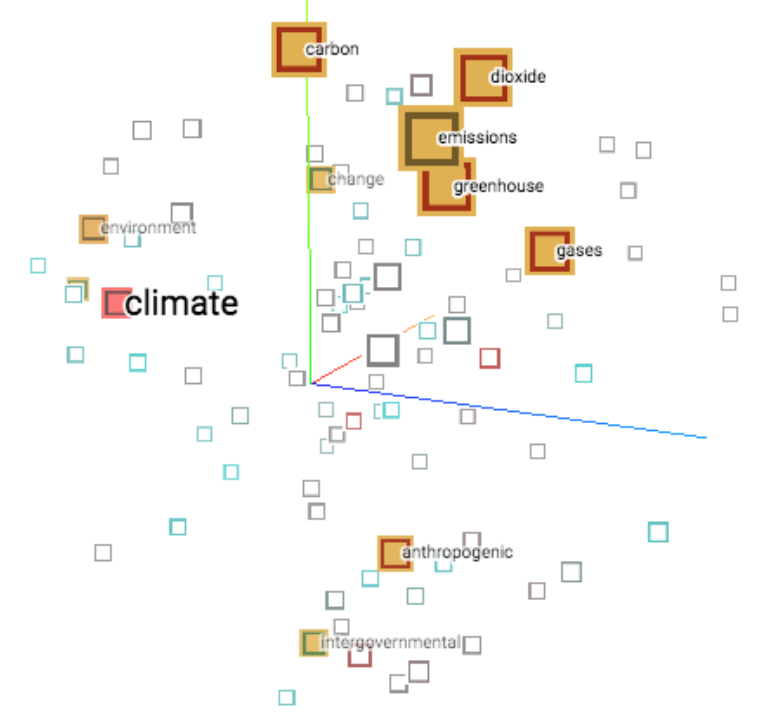}
\label{fig:kyoto}
\end{figure}

\begin{figure}[h]
\caption{Closest words to "climate" in the Paris agreement (PCA projection from Tensorboard). }
\centering
\includegraphics[width=0.47\textwidth]{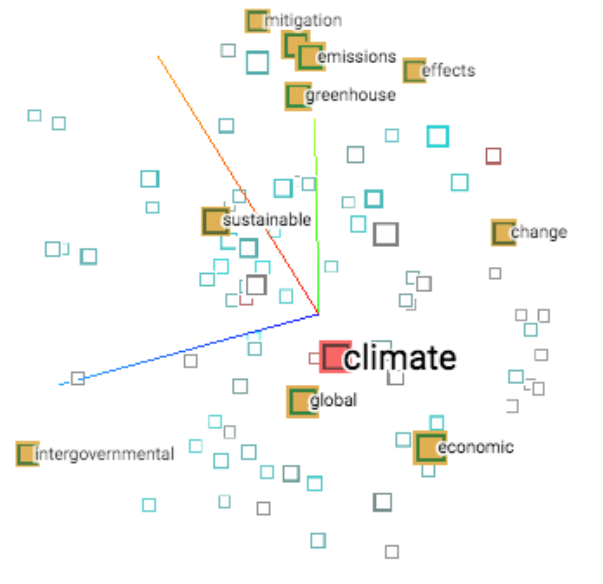}
\label{fig:paris}
\end{figure}

Let us look, for instance, at key terms used in the Kyoto protocol (8483 words text document, see figure \ref{fig:kyoto}) and the Paris agreement (7383 words text document, see figure \ref{fig:paris}), and highlight the closest words to "climate".

Below are few of their characteristics.
\begin{itemize}
\item The reader will not be surprised to find neighboring terms such as "sustainable", "greenhouse", "global", and "economic".
  \item The two texts have a neutral sentiment (grey word boxes), with perhaps a slightly more positive tone in the Paris agreement (blue word boxes), which is what one would expect for international agreements.
    \item Four out of ten terms appear in both word clouds. The former cloud puts the emphasis on change and intergovernmental actions, and the latter on economics and sustainability. But they remain consistent, which is remarkable given that they are based on completely different texts.
\end{itemize}

In the case of the 2016 presidential election, IBM Watson tells us that the most relevant words from Donald Trump were trade deals (expressed with sadness), ISIS (anger), and Clinton (disgusted). Hillary Clinton was very much focused on new jobs and the affordable care act. \par

\begin{figure}[h]
\caption{Donald Trump’s words that related to trade deals (U.S. election televised debates).}
\centering
\includegraphics[width=0.4\textwidth]{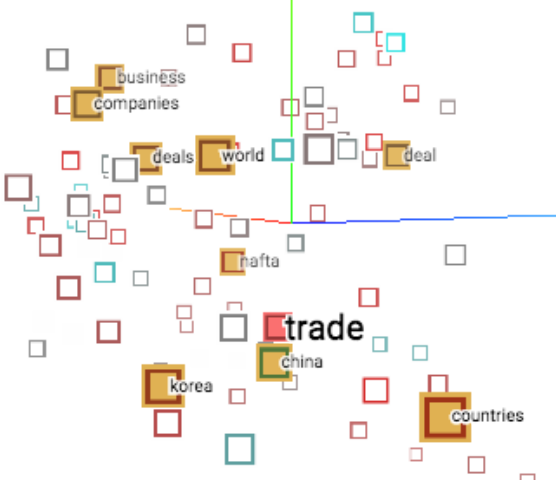}
\label{fig:trade-trump-2016}
\end{figure}

Thanks to political footprints, we can determine that words most closely related to trade (deals) in Donald Trump’s vocabulary were Nafta (used with fear), China, Korea, the world (used with sadness), countries (disgust), and other economic terms such as tax, business, and companies (see figure \ref{fig:trade-trump-2016}).\par

\begin{figure}[h]
\caption{Hillary Clinton’s words that related to the affordable care act (U.S. election televised debates). }
\centering
\includegraphics[width=0.38\textwidth]{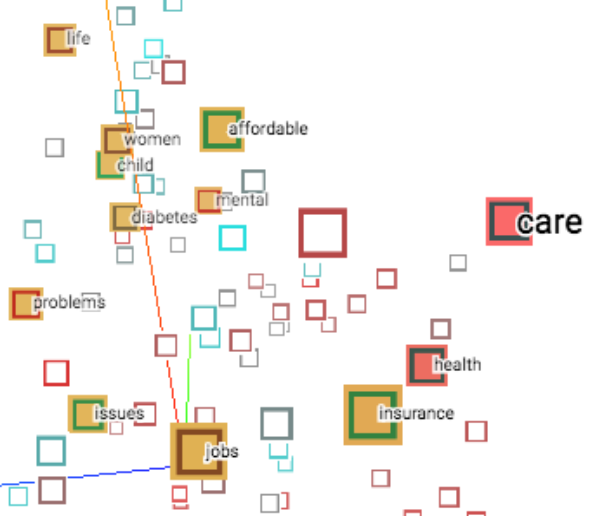}
\label{fig:care-clinton-2016}
\end{figure}

Hillary Clinton’s care-related vocabulary included terms such as women (used with disgust, probably in the context of gender equality), children, insurance and health, both presented positively (see figure \ref{fig:care-clinton-2016}).\par

What makes us confident about our results is the appearance of terms that were widely commented on during the 2016 U.S. election: "women" and "health care" for Hillary Clinton, "China" and "Nafta" for Donald Trump.\par

What is important to understand, and what is at the core of political footprints, is that word similarities have been identified without any human intervention: they have been discovered by machines based on how frequently words appear together, either on Wikipedia or other large corpora of text. This is what allows us to produce a more objective analysis of political discourses. To be clear, there is a strong cultural bias, originating in how these words have been used on Wikipedia, news feeds, etc. But it is not coming from the researcher performing the analysis.\par
There is no such thing as objective political discourse analysis. However, solutions to reduce bias from sources such as Wikipedia could be to train VSMs on a set of data coming from different sources (Wikipedia, Google News, social media, political discourses themselves, etc.), or to agree on the criteria for selecting unbiased texts. We are highly skeptical about the latter: a paradigm at the core of machine learning is that a large set of data is often closer to reality than a small curated set.\par
Word similarities are not inferred from the discourse we analyze but from the large corpus of text that was used to train words. Comparing distances (cosine similarity) between words does not provide any information about the discourse we study, but only about the culture and language it is based on; information about the discourse lies in the choice of these words instead of others, their relevance and associated emotion.\par
This heuristic was performed manually but could have been automated using unsupervised machine learning techniques such as k-means~\cite{kmeans}. It would have then been possible to compare their standard implementation with one that takes into account word relevance. Using the most relevant words to define our word clusters makes sense intuitively since they are “at the center” of the discourse.
\subsection{Heuristic 2: compare how a theme is appropriated by each debate participant}

In this heuristic, we choose a term or theme of our choice. Let us assume we are interested in American "values".\par
We select the 20 closest words to "values" (i.e. "social", "civilization", "inequality", "liberty", etc.) and see how many presidential candidates have used them.\par
We then choose a couple of these terms, ideally those that have been used by many participants and that have many different meanings. "Social" and "interest" were picked in our 2016 US election example, but it could also have been “ethical”, “principle” or “freedom”: any term that can be used differently depending on a participant's political views.\par
We finally compute the list of related terms for each candidate using our first heuristic.\par

\begin{figure}[h]
\caption{Value-related terms used during the 2008 U.S. presidential election (televised debates).}
\centering
\includegraphics[width=0.47\textwidth]{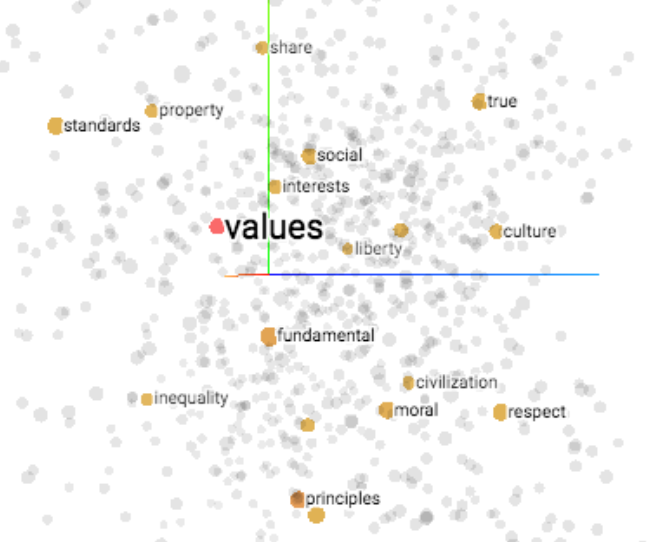}
\label{fig:values-2008}
\end{figure}

Figure \ref{fig:values-2008} shows the main value-related terms used during the 2008 U.S. presidential election. As we can see, debated values were social, related to property, civilization, inequality, and liberty.\par

\begin{figure}[h]
\caption{Value-related terms used during the 2016 U.S. presidential election (televised debates).}
\centering
\includegraphics[width=0.4\textwidth]{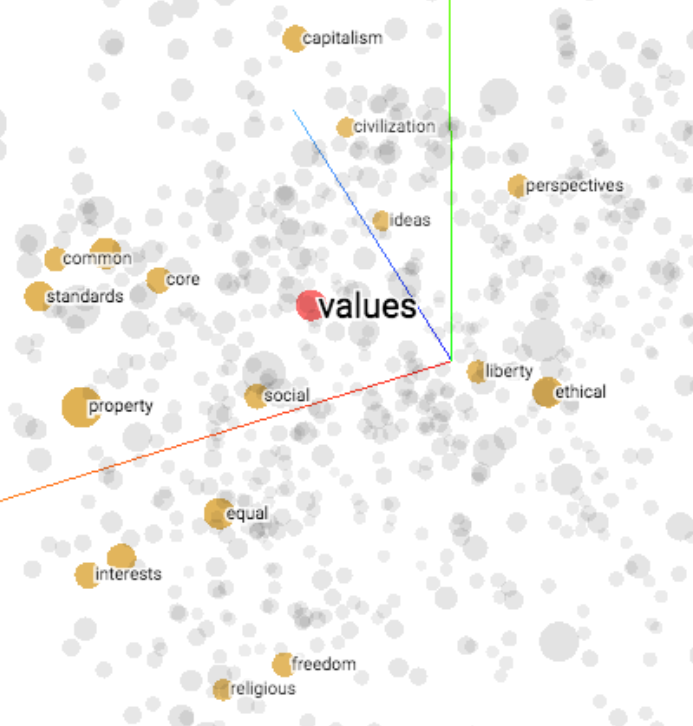}
\label{fig:values-2016}
\end{figure}

Civilization, property, standards, social, interests, and liberty also appear during the 2016 U.S. presidential election (see figure \ref{fig:values-2016}). This is an indication of both political footprints' robustness and the consistency of the terms used from one election to the other.\par

We have, with the two word clouds above, the confirmation that social values played an important role in the two elections. Let us now see how three candidates have referred to this topic (social-related terms in the GloVe vector space model).\par

\begin{figure}[h]
\caption{John McCain’s topics that related to social matters (2008 primaries debates).}
\centering
\includegraphics[width=0.4\textwidth]{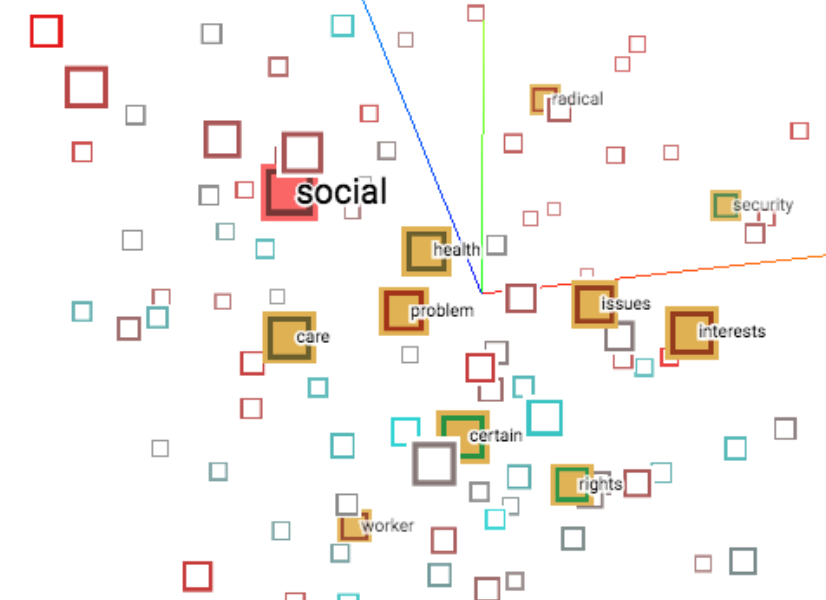}
\label{fig:mccain-social-2008}
\end{figure}

When John McCain talked about social issues during the 2008 Republican primaries, he talked mostly about healthcare, security (“radical” was used with fear), jobs (“worker” generated some anger), and rights. He decided to not focus on a single topic and instead discussed various social issues (see figure \ref{fig:mccain-social-2008}).\par

\begin{figure}[h]
\caption{Barack Obama’s topics that related to social issues (2008 primaries debates).}
\centering
\includegraphics[width=0.4\textwidth]{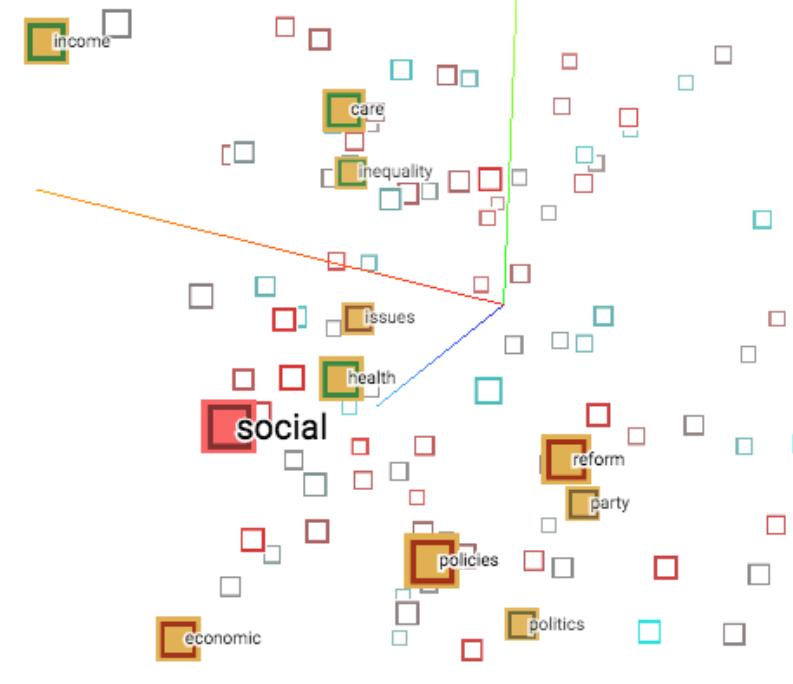}
\label{fig:obama-social-2008}
\end{figure}

During the 2008 Democrat primaries, Barrack Obama decided to focus more on health care. Jobs were high on his agenda (“income”, “economic”), and inequalities were an important topic as well (see figure \ref{fig:obama-social-2008}).\par

\begin{figure}[h]
\caption{Bernie Sanders’ topics that related to social matters (2016 primaries debates).}
\centering
\includegraphics[width=0.4\textwidth]{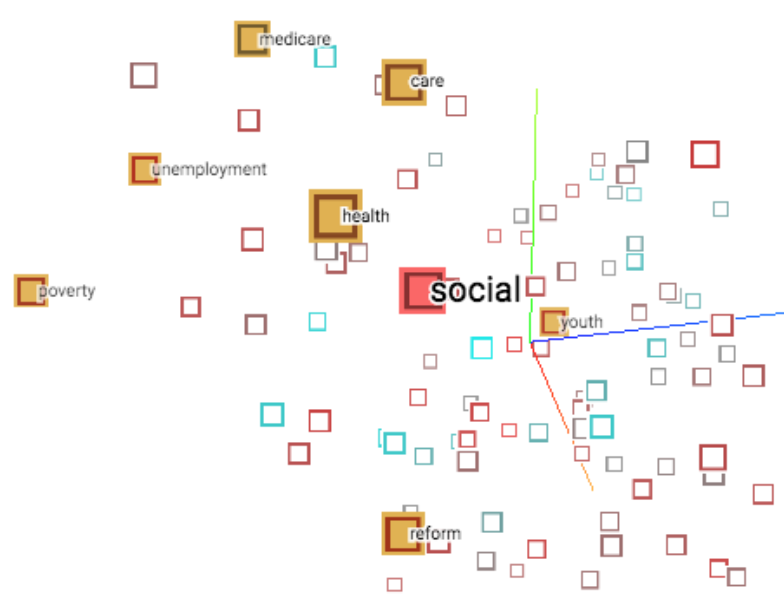}
\label{fig:social-bernie-2016}
\end{figure}

In 2016, Bernie Sanders also put health care at the top of his social agenda. His stance on economic issues centered more on the poor (income, unemployment, poverty) than the two other candidates, see figure \ref{fig:social-bernie-2016}.\par

\begin{figure}[h]
\caption{Barack Obama’s topics that related to American interests (2008 primaries debates).}
\centering
\includegraphics[width=0.38\textwidth]{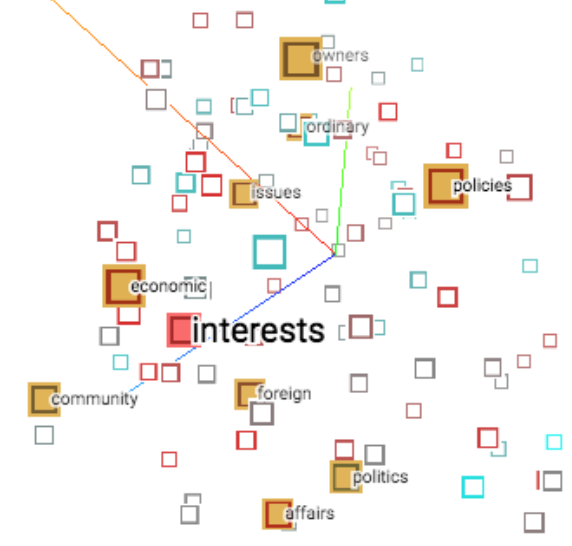}
\label{fig:obama-interests-2008}
\end{figure}

Let us explore what American “interests” meant during the two presidential elections, and look at related words from Barack Obama and Donald Trump.
During the 2008 Democrat primary, Barack Obama’s closest topics to (American) interests were foreign affairs (with a sense of fear), the economy (“owners”,” economic”), and social issues (“community”, “ordinary”, “social”) (see figure \ref{fig:obama-interests-2008}).\par

\begin{figure}[h]
\caption{Donal Trump’s topics that related to American interests (2008 primaries debates).}
\centering
\includegraphics[width=0.47\textwidth]{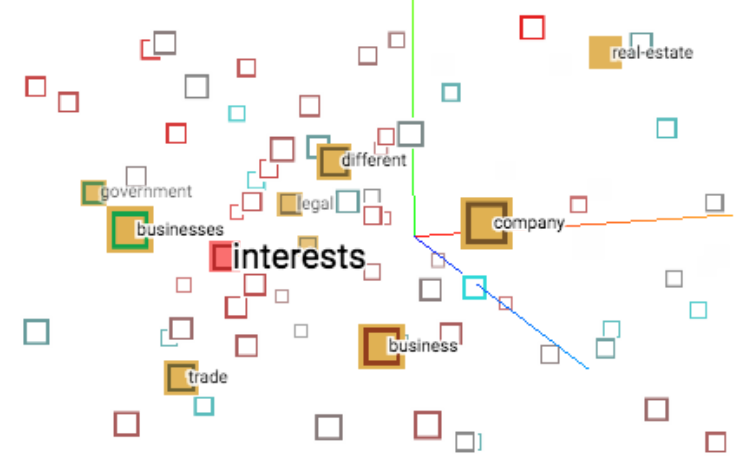}
\label{fig:interests-trump-2016}
\end{figure}

Donald Trump had something different in mind when talking about American interests (2016 Republican primaries). His political footprint encompasses matters centering on the economy and trade deals, see figure \ref{fig:interests-trump-2016}.\par

Using a unique reference model is both the biggest weakness and strength of our political footprints. On the one hand, it does not accurately represent the words’ meaning during each campaign (election years were not the same as the years covered by our VSM), but on the other hand it provides an easy way to compare them: the semantic relation between words have been defined at the GloVe level, we can thus compare texts that have not necessarily used the same words to address the same issue.

\subsection{Heuristic 3: sort discourses by style and affinity}

The purpose of this last section is to test whether the general properties of our political footprints could be used for political discourse analysis. For this, we visualized political footprints coloured by relevance, sentiment, and emotions. We calculated their center of gravity, or centroids~\cite{centroid}, and compared their distance with one another.\par
None of these approaches were, at least in our current implementation, conclusive. A few interesting properties were revealed in our US election example, but not enough to exclude them being mere coincidences. Bernie Sanders' political footprint was more focused than the others: Wall Street was detected as being by far the most relevant of his topics (along with Hillary Clinton). But this might also have been due to a strategy of repeating the same words instead of describing his views in different ways. We cannot conclude that Bernie Sanders was fundamentally more focused.\par
Hillary Clinton's emotions were less visible. But as explained above, this might have been due to her sarcastic tone, and more subtle ways of expressing her emotions.\par
Finally, except possibly for the fact that Bernie Sanders' centroid was the furthest from Donald Trump’s, their positions did not correspond to any of our intuitions. \par
Using centroids is arguably a simplistic way to look at political footprints: they do not take into account relevance, sentiments and emotions. For instance, seeing Islam as a positive and secondary topic counts exactly the same in our model as seeing Islam as a negative and central topic.

\section{Conclusions}

In this paper, we have presented a very simple implementation of political footprints: one that can be computed on any personal computer and still leverage some of the semantic value embedded in vector space models. Identifying the “real” meaning of a political discourse is in itself an impossible endeavor. However, political footprints fit surprisingly well with the discourse of each US presidential candidate and how they have been covered in the press. They have been obtained with nearly no human intervention, which makes them a potentially very useful tool for political discourse analysis.\par
A better way to test their validity would be to compare resulting word clouds with those that the public, the commentators, and the authors of a discourse themselves would draw. We could ask an audience to recognize a discourse based on its political footprint, and measure the success rate. Or we could compare political footprints of the same politician in different contexts (debates, rallies, public declarations, twitter) and test how consistent they are.\par
We have suggested various improvements, such as using a domain-specific and open source solution instead of IBM Watson, using a pre-trained vector space model that does not break compound nouns, and combining existing clustering techniques with our heuristics. \par
Political footprints are meant to be used by anyone interested in political discourse analysis, in highlighting some of the different meanings a word can have in politics, and the underlying semantic tensions underlying any political debate. 
It is hoped that such tools will help researchers and commentators alike reduce their dependence on personal opinions when analyzing political discourse. Political footprints emphasize what politicians have control over and responsibility for: their own words.

"When a man commits himself to anything, fully realising that he is not only choosing what he will be, but is thereby at the same time a legislator deciding for the whole of mankind – in such a moment a man cannot escape from the sense of complete and profound responsibility"~\cite{sartre}.

\section{Acknowledgement}
We would like to thank Niel Chah~\cite{niel} for his support during the political footprint development process and Araz Taeihagh~\cite{araz} for his feedback and time spent reviewing the paper.

\bibliography{thebibliography}

\end{document}